\documentclass[10pt,times,mathptm,psfig,twocolumn,journals]{IEEEtran}
\usepackage{url}
\usepackage{amssymb}
\usepackage{array}
\usepackage{amsmath}
\usepackage{graphicx}
\usepackage{booktabs}
\usepackage{epsfig}
\usepackage{ulem}
\usepackage{hyperref}
\hypersetup{pdfborder={0 0 0}, colorlinks=true, linkcolor=red, citecolor=blue}
\usepackage{subfigure}
\usepackage{cite}
\usepackage{epstopdf}
\usepackage{amsmath}
\usepackage[misc]{ifsym} 
\usepackage{bm}
\usepackage{mathrsfs}
\usepackage{algorithm}
\usepackage{algorithmic}
\begin{document}

\title{A Deep Ordinal Distortion Estimation Approach for Distortion Rectification}

\author{Kang~Liao, Chunyu~Lin, Yao~Zhao
\thanks{Kang Liao, Chunyu Lin, Yao Zhao are with the Institute of Information Science, Beijing Jiaotong University, Beijing 100044, China, and also with the Beijing Key Laboratory of Advanced Information Science and Network Technology, Beijing 100044, China (email: kang\_liao@bjtu.edu.cn, cylin@bjtu.edu.cn, yzhao@bjtu.edu.cn. Corresponding author: Chunyu Lin.)}
}

\maketitle
\graphicspath{{img/}}
\begin{abstract}
Radial distortion has widely existed in the images captured by popular wide-angle cameras and fisheye cameras. Despite the long history of distortion rectification, accurately estimating the distortion parameters from a single distorted image is still challenging. The main reason is that these parameters are implicit to image features, influencing the networks to learn the distortion information fully. In this work, we propose a novel distortion rectification approach that can obtain more accurate parameters with higher efficiency. Our key insight is that distortion rectification can be cast as a problem of learning an \textit{ordinal distortion} from a single distorted image. To solve this problem, we design a local-global associated estimation network that learns the ordinal distortion to approximate the realistic distortion distribution. In contrast to the implicit distortion parameters, the proposed ordinal distortion has a more explicit relationship with image features, and significantly boosts the distortion perception of neural networks. Considering the redundancy of distortion information, our approach only uses a patch of the distorted image for the ordinal distortion estimation, showing promising applications in efficient distortion rectification. In the distortion rectification field, we are the first to unify the heterogeneous distortion parameters into a learning-friendly intermediate representation through ordinal distortion, bridging the gap between image feature and distortion rectification. The experimental results demonstrate that our approach outperforms the state-of-the-art methods by a significant margin, with approximately 23\% improvement on the quantitative evaluation while displaying the best performance on visual appearance. The code is available at \url{https://github.com/KangLiao929/OrdinalDistortion}.
\end{abstract}

\IEEEpeerreviewmaketitle

\section{Introduction}
\IEEEPARstart {I}{MAGES} captured by wide-angle camera usually suffer from a strong distortion, which influences the important scene perception tasks such as the object detection and recognition \cite{ref49, ref50, ref57}, semantic segmentation \cite{ref51, ref52}, and image denoising \cite{ref58, ref59}. The distortion rectification tries to recover the real geometric attributes from distorted scenes. It is a fundamental and indispensable part of image processing, which has a long research history extending back 60 years. In recent, distortion rectification through deep learning has attracted increasing attention\cite{ref10, ref11, ref47, ref44, ref46, ref48, ref55}. 

Accurately estimating the distortion parameters derived from a specific camera, is a crucial step in distortion rectification. However, two main limitations that make the distortion parameters learning challenging. (i) The distortion parameters are not observable and hard to learn from a single distorted image, such as the principal point and distortion coefficients. Compared with the intuitive targets, such as the object classification and bounding box detection studied in other research regions, the distortion parameters have a more complicated and implicit relationship with image features. As a result, the neural networks obtain an ambiguous and insufficient distortion perception, which leads to inaccurate estimation and poor rectification performance. (ii) The different components of distortion parameters have different magnitudes and ranges of values, showing various effects on an image's global distortion distribution. Such a heterogeneous representation confuses the distortion cognition of neural networks and causes a heavy imbalance problem during the training process. 

To overcome the above limitations, previous methods exploit more guided features such as the semantic information and distorted lines \cite{ref11, ref47}, or introduce the pixel-wise reconstruction loss \cite{ref44, ref46, ref48}. However, the extra features and supervisions impose increased memory/computation cost. In this work, we would like to draw attention from the traditional calibration objective to a learning-friendly perceptual target. The target is to unify the implicit and heterogeneous parameters into an intermediate representation, thus bridging the gap between image feature and distortion estimation in the field of distortion rectification.

\begin{figure*}[t]
 \centering
  \includegraphics[width=0.85\linewidth]{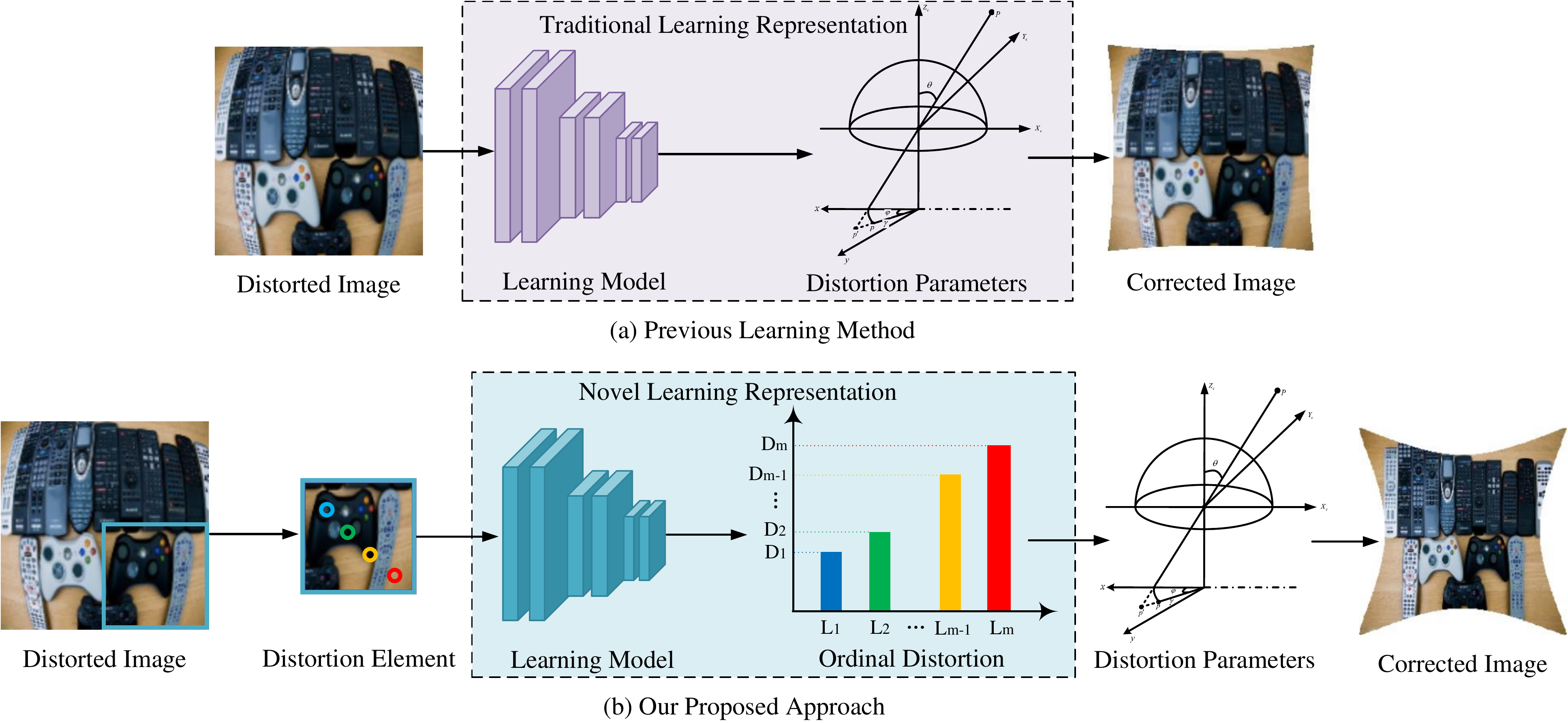}
  \caption{Method Comparisons. (a) Previous learning methods, (b) Our proposed approach. We aim to transfer the traditional calibration objective into a learning-friendly representation. Previous methods roughly feed the whole distorted image into their learning models and directly estimate the implicit and heterogeneous distortion parameters. In contrast, our proposed approach only requires a part of a distorted image (distortion element) and estimates the ordinal distortion. Due to its explicit description and homogeneity, we can obtain more accurate distortion estimation and achieve better corrected results.}
		\label{Fig:1}
\end{figure*}

In particular, we redesign the whole pipeline of deep distortion rectification and present an intermediate representation based on the distortion parameters. The comparison of the previous methods and the proposed approach is illustrated in Fig. \ref{Fig:1}. Our key insight is that distortion rectification can be cast as a problem of learning an \textit{ordinal distortion} from a distorted image. The ordinal distortion indicates the distortion levels of a series of pixels, which extend outward from the principal point. To predict the ordinal distortion, we design a local-global associated estimation network optimized with an ordinal distortion loss function. A distortion-aware perception layer is exploited to boost the feature extraction of different degrees of distortion.

The proposed learning representation offers three unique advantages. First, the ordinal distortion is directly perceivable from a distorted image, and it solves a more straightforward estimation problem than the implicit metric regression. As we can observe, the farther the pixel is away from the principal point, the larger the distortion degree is, and vice versa. This prior knowledge enables the neural networks to build a clear cognition with respect to the distortion distribution. Thus, the learning model gains a sufficient distortion perception of image features and shows faster convergence, without any extra feature guidances and pixel-wise supervisions.  

Second, the ordinal distortion is homogeneous as all its elements share a similar magnitude and description. Therefore, the imbalanced optimization problem no longer exists during the training process, and we do not need to focus on the cumbersome factor-balancing task anymore. Compared to the distortion parameters with different types of components, our learning model only needs to consider one optimization objective, thus achieving more accurate estimation and more realistic rectification results.

Third, the ordinal distortion can be estimated using only a part of a distorted image. Unlike the semantic information, the distortion information is redundant in images, showing the central symmetry and mirror symmetry to the principal point. Consequently, the efficiency of rectification algorithms can be significantly improved when taking the ordinal distortion estimation as a learning target. More importantly, the ordinal relationships are invariant to monotonic transformations of distorted images, thereby increasing the robustness of the rectification algorithm. 

With lots of experiments, we verify that the proposed ordinal distortion is more suitable than the distortion parameters as a learning representation for deep distortion rectification. The experimental results also show that our approach outperforms the state-of-the-art methods with a large margin, approximately 23\% improvement on the quantitative evaluation while using fewer input images, demonstrating its efficiency on distortion rectification.

The rest of this paper is organized as follows. We first introduce the related work in Section \ref{sec2}. We then present our approach in Section \ref{sec3}. The experiments are provided in Section \ref{sec4}. Finally, we conclude this paper in Section \ref{sec5}. 

\section{Related Work}
\label{sec2}
In this section, we briefly review the previous distortion rectification methods and classify them into two groups: the traditional vision-based one and the deep learning one.

\subsection{Traditional Distortion Rectification} 
There is a rich history of exploration in the field of distortion rectification. The most common method is based on a specific physical model. \cite{63, 64, 65} utilized a camera to capture several views of a 2D calibration pattern that covered points, corners, or other features, and then computed the distortion parameters of the camera. However, these methods cannot handle images captured by other cameras and thus are restricted to the application scenario. Self-calibration was leveraged for distortion parameter estimation in \cite{ref4, ref5, ref6}; however, the authors failed in the geometry recovery using only a single image. To overcome the above limitations and achieve automatic distortion rectification, Bukhari et al. \cite{ref7} employed a one-parameter camera model \cite{ref8} and estimated distortion parameters using the detected circular arcs. Similarly, \cite{ref9, ref45} also utilized the simplified camera model to correct the radial distortion in images. However, these methods perform poorly on scenes that are lacking enough hand-crafted features. Thus, the above traditional methods are difficult to handle on the single distorted image rectification in various scenes.
\begin{figure*}
	\begin{center}
		\includegraphics[width=1\linewidth]{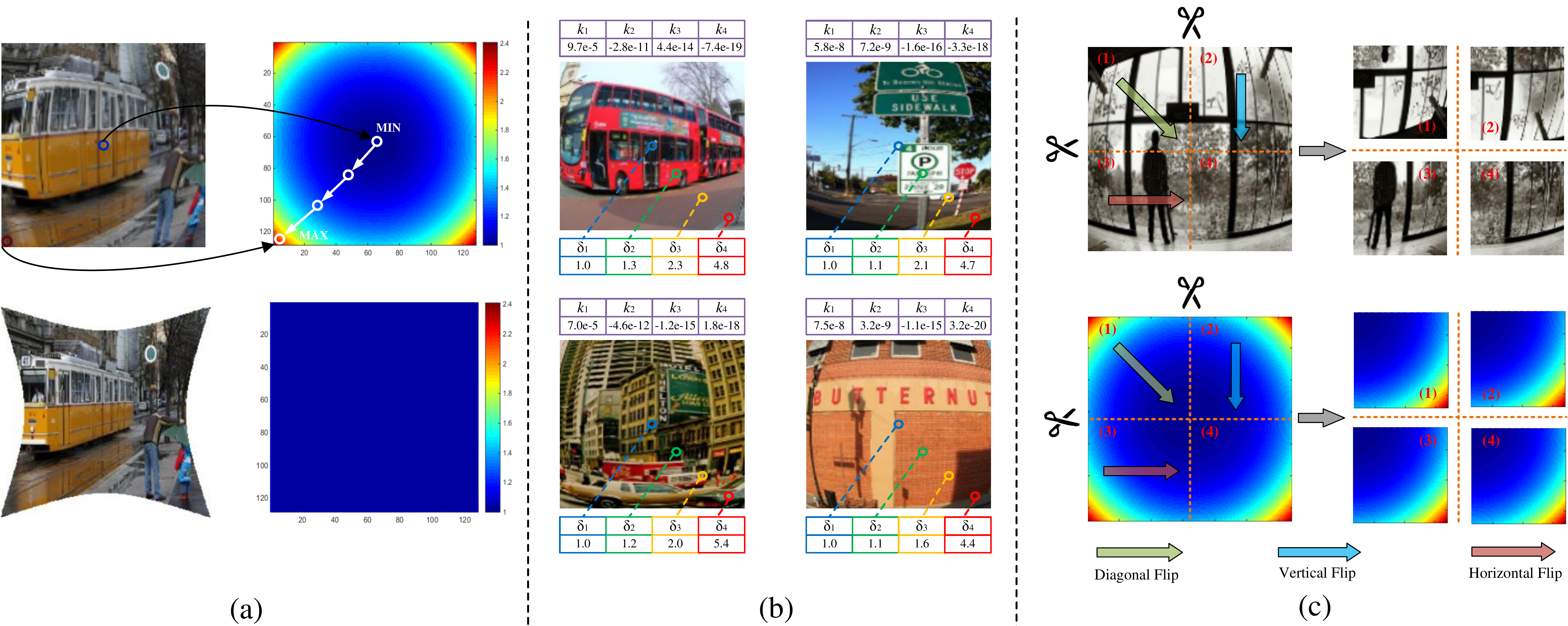}
		\caption{Attributes of the proposed ordinal distortion. (a) Explicitness. The ordinal distortion is observable in an image and explicit to image features, which describes a series of distortion levels from small to large (top); the ordinal distortion always equals one in an undistorted image (bottom). (b) Homogeneity. Compared with the heterogeneous distortion parameters $\mathcal{K} = [k_1 \ \ k_2 \ \ k_3 \ \ k_4]$, the ordinal distortion $\mathcal{D} = [\delta_1 \ \ \delta_2 \ \ \delta_3 \ \ \delta_4]$ is homogeneous, representing the same concept of distortion distribution. (c) Redundancy. After different flip operations, although the semantic features of four patches have not any relevance (top), the ordinal distortion of four patches keeps the same in distribution with each other (bottom).}
		\label{Fig:2}
	\end{center}
	%\vspace{-0.5em}
\end{figure*}
\subsection{Deep Distortion Rectification} 
In contrast to the long history of traditional distortion rectification, learning methods began to study distortion rectification in the last few years. Rong et al. \cite{ref10} quantized the values of the distortion parameter to 401 categories based on the one-parameter camera model \cite{ref8} and then trained a network to classify the distorted image. This method achieved the deep distortion rectification for the first time, while the coarse values of parameters and the simplified camera model severely influenced its generalization ability. To expand the application, Yin et al. \cite{ref11} rectified the distortion in terms of the fisheye camera model using a multi-context collaborative deep network. However, their correction results heavily rely on the semantic segmentation results, leading to a strong cascading effect. Xue et al. \cite{ref47} improved the performance of distortion parameter estimation by distorted lines. In analogy to traditional methods \cite{ref7, ref9, ref45}, the extra introduced hand-crafted features limit the robustness of this algorithm and decrease the efficiency of the rectification. Note that the above methods directly estimates distortion parameters from a single distorted image, such an implicit and heterogeneous calibration objective hinders sufficient learning concerning the distortion information. To solve the imbalance problem in the distortion parameter estimation, recent works \cite{ref44, ref46, ref48} optimized the image reconstruction loss rather than the parameters regression loss for rectification. However, their models are based on the parameter-free mechanism and cannot estimate the distortion parameters, which are important for the structure from motion and camera calibration. Manuel et al. \cite{ref55} proposed a parameterization scheme for the extrinsic and intrinsic camera parameters, but they only considered one distortion coefficient for the rectification and cannot apply the algorithm to more complicated camera models.

With the proposed intermediate representation, i.e., ordinal distortion, our approach can boost the efficient learning of neural networks and eliminate the imbalance problem, obtaining accurate parameters for better rectification performance.

\begin{figure*}
	\begin{center}
		\includegraphics[width=1\linewidth]{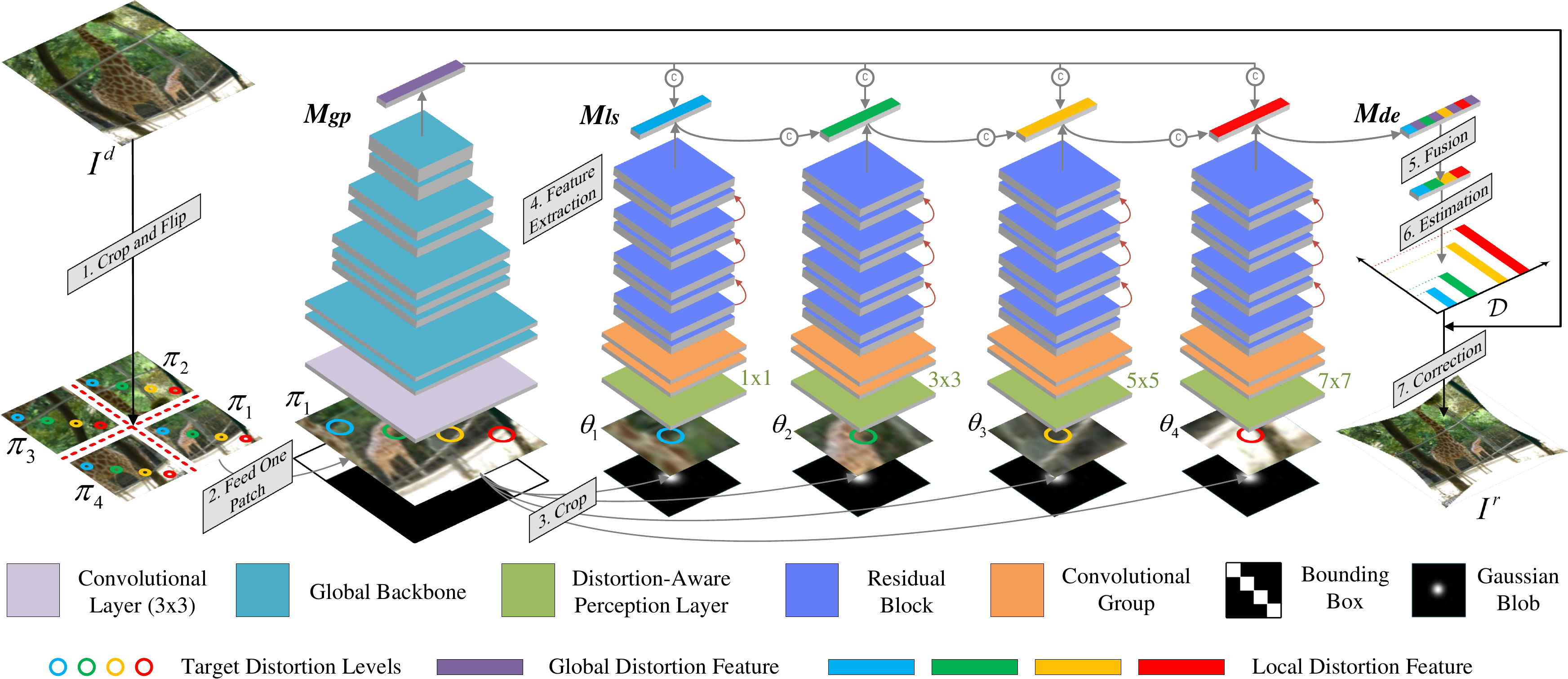}
		\caption{Architecture of the proposed network. This network consists of a global perception module $M_{gp}$, local Siamese module $M_{ls}$, and distortion estimation module $M_{de}$. During the network's training process, we use four parts, i.e., distortion elements: $\Pi = [\pi_1 \ \ \pi_2 \ \ \pi_3 \ \ \pi_4]$ of the distorted image $I^d$ to train its ability of distortion perception sequentially, in which the distortion blocks: $\Theta = [\theta_1 \ \ \theta_2  \ \ \theta_3 \ \ \theta_4]$ derived from a distortion element provide the local distortion information. In the test or application stage, we only need one part of the input distorted image to estimate the ordinal distortion $\mathcal{D}$. Finally, the rectified image $I^r$ can be generated using the estimated ordinal distortion and the input distorted image.}
		\label{Fig:3}
	\end{center}
	%\vspace{-0.5em}
\end{figure*}

\section{Approach}
\label{sec3}
In this section, we describe how to learn the ordinal distortion given a single distorted image. We first define the proposed objective in Section \ref{s31}. Next, we introduce the network architecture and training loss in Section \ref{s32}. Finally, Section \ref{s33} describes the transformation between the ordinal distortion and distortion parameter.

\subsection{Problem Definition}
\label{s31}
\subsubsection{Parameterized Camera Model}
We assume that a point in the distorted image is expressed as $\mathbf{P} = [x, y]^{\rm T} \in {\mathbb{R}}^{2}$ and a corresponding point in the corrected image is expressed as $\mathbf{P'} = [x', y']^{\rm T} \in {\mathbb{R}}^{2}$. The polynomial camera model can be described as
\begin{equation}\label{eq1}
\begin{split}
&x' = x(1 + {k_1}{{r}^2} + {k_2}{{r}^4} + {k_3}{{r}^6} + {k_4}{{r}^8} + \cdots) \\
&y' = y(1 + {k_1}{{r}^2} + {k_2}{{r}^4} + {k_3}{{r}^6} + {k_4}{{r}^8} + \cdots), \\
\end{split}
\end{equation}
where $[k_1\ \ k_2 \ \ k_3 \ \ k_4 \ \ \cdots]$ are the distortion coefficients, $r$ is the Euclidean distance between the point $\mathbf{P}$ and the principal point $\mathbf{C} = [x_c, y_c]^{\rm T}$ in the distorted image, which can be expressed as
\begin{equation}\label{eq2}
r = \sqrt{(x - x_c)^2 + (y - y_c)^2}.
\end{equation}

This polynomial camera model fits well for small distortions but requires more distortion parameters for severe distortions. As an alternative camera model, the division model is formed by:
\begin{equation}\label{eq3}
\begin{split}
&x' = \frac{x}{1 + {k_1}{{r}^2} + {k_2}{{r}^4} + {k_3}{{r}^6} + {k_4}{{r}^8} + \cdots} \\
&y' = \frac{y}{1 + {k_1}{{r}^2} + {k_2}{{r}^4} + {k_3}{{r}^6} + {k_4}{{r}^8} + \cdots}.\\
\end{split}
\end{equation}

Compared with the polynomial camera model, the division model requires fewer parameters in terms of the strong distortion and is thus more suitable for approximating wide-angle cameras \cite{60}.

\subsubsection{Ordinal Distortion}
\label{sec3.2}
As mentioned above, most previous learning methods correct the distorted image based on the distortion parameters estimation. However, due to the implicit and heterogeneous representation, the neural network suffers from the insufficient learning problem and imbalance regression problem. These problems seriously limit the learning ability of neural networks and cause inferior distortion rectification results. To address the above problems, we propose a fully novel concept, i.e., ordinal distortion. Fig. \ref{Fig:2} illustrates the attributes of the proposed ordinal distortion.

The ordinal distortion represents the image feature in terms of the distortion distribution, which is jointly determined by the distortion parameters and location information. We assume that the camera model is the division model, and the ordinal distortion $\mathcal{D}$ can be defined as
\begin{equation}\label{eqd}
\begin{split}
\mathcal{D} &= [\delta(r_1) \ \ \delta(r_2) \ \ \delta(r_3) \ \  \cdots \ \ \delta(r_n)], \\
&0 \leq r_1 < r_2 < r_3 < \cdots < r_n \leq R,\\
\end{split}
\end{equation}
where $R$ is the maximum distance between a point and the principal point, $\delta(\cdot)$ indicates the distortion level of a point $P_i$ in the distorted image:

\begin{equation}\label{eq5}
\delta(r_i) = \frac{x_i}{x'_i} = \frac{y_i}{y'_i} = 1 + {k_1}{{r_i}^2} + {k_2}{{r_i}^4} + {k_3}{{r_i}^6} + {k_4}{{r_i}^8} + \cdots.
\end{equation}

Intuitively, the distortion level expresses the ratio between the coordinates of $\mathbf{P}$ and $\mathbf{P'}$. The larger the distortion level is, the stronger the distortion of a pixel is, and vice versa. For an undistorted or ideally rectified image, $\delta(\cdot)$ always equals 1. Therefore, the ordinal distortion represents the distortion levels of pixels in a distorted image, which increases outward from the principal point sequentially.

We assume the width and height of a distorted image are $W$ and $H$, respectively. Then, the distortion level satisfies the following equation:
\begin{equation}\label{eq7}
\begin{split}
\delta(x_i, y_i) &= \delta(\frac{W}{2} - x_i + x_c, y_i) = \delta(x_i, \frac{H}{2} - y_i + y_c) \\
&= \delta(\frac{W}{2} - x_i + x_c, \frac{H}{2} - y_i + y_c).\\
\end{split}
\end{equation}
Thus, the ordinal distortion displays the mirror symmetry and central symmetry to the principal point in a distorted image. This prior knowledge ensures less data required in the ordinal distortion estimation process. 

\subsection{Network}
\label{s32}
Our network consists of three main modules: global perception module $M_{gp}$, local Siamese module $M_{ls}$, and distortion estimation module $M_{de}$ as shown in Fig. \ref{Fig:3}. The first module extracts the global distortion features from a patch of the input distorted image. The second module extracts the local distortion features from a series of distortion blocks, corresponding to the different distortion levels. The final one fuses the extracted global and local distortion features and estimates the proposed ordinal distortion.

\subsubsection{Network Input}
The network input includes two parts. The first is the global distortion context, which provides a distortion element $\pi_i \in \Pi$ with the overall distortion information. The second is the local distortion context, which provides the distortion blocks $\Theta = [\theta_1 \ \ \theta_2  \ \ \theta_3 \ \ \cdots \ \ \theta_n]$ with the detailed distortion levels. Considering the principal point is slightly disturbed in the image center, we first cut the distorted image into four patches along the center of the image, and dub these patches as distortion elements $\Pi = [\pi_1 \ \ \pi_2 \ \ \pi_3 \ \ \pi_4]$ with size of $\frac{W}{2}\times\frac{H}{2}\times3$. Although most distortion information covers in one patch, the distortion distribution of each patch is spatially different. To normalize this diversity, we flip three of the four elements to keep a similar distortion distribution with that of the selected one. As shown in Fig. \ref{Fig:2} (c), the top left, top right, and bottom left distortion parts are handled with the diagonal, vertical, and horizontal flip operations, respectively.

\begin{figure}
	\begin{center}
		\includegraphics[width=.7\linewidth]{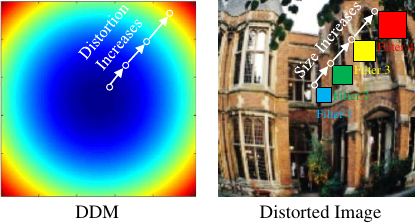}
		\caption{Motivation of the designed distortion-aware perception layer. Left: the distortion distribution map (DDM) that describes the degree of distortion for each pixel. Right: the corresponding distorted image. Particularly, we use the filters with increasing sizes to perceive the increasing degrees of distortions along the extended path (the white arrows).}
		\label{Fig:da}
	\end{center}
	%\vspace{-0.5em}
\end{figure}

We further crop a distortion element into the distortion blocks $\Theta = [\theta_1 \ \ \theta_2  \ \ \theta_3 \ \ \cdots \ \ \theta_n]$, in which a block $\theta_i$ is leveraged to provide the local distortion feature and predict the distortion level $\delta_i$. To boost neural networks to learn the distortion features, we construct the masks consisting of the bounding boxes $\mathcal{M}_B \in {\mathbb{R}}^{w_b \times h_b \times 1}$ and Gaussian blobs $\mathcal{M}_G \in {\mathbb{R}}^{w_g \times h_g \times 1}$ of the distortion blocks, where $w_b = \frac{W}{2}, h_b = \frac{H}{2}, w_g = \frac{W}{2n}, h_g = \frac{H}{2n}$. Concretely, the mask represents the Region of Interest (RoI) of input data, which offers the ranges of global and local distortion information for $\mathcal{M}_B$ and $\mathcal{M}_G$:

\begin{eqnarray}
\label{eq-mb}
\mathcal{M}_{B}(p, q)=
\left\{
\begin{array}{lll}
255, \ \ \ \ &if \ \  (p, q) \in \Omega. \\
0, &otherwise.
\end{array}
\right.
\end{eqnarray}
where $\Omega = \{(p, q)|(t-1)w_g \le p \le t w_g, (t-1)h_g \le q \le t h_g\}$ is the region of distortion blocks and $t \in \mathbb{Z}$ derives from the range of $[1, n]$. For $\mathcal{M}_G$, the mask value can be weighted by the Gaussian Distribution $X \sim  \mathcal{N}(\mu,\,\sigma^{2})$:
\begin{equation}\label{eq-mg}
\mathcal{M}_{G}(d) = 255\frac{1}{\sigma\sqrt{2\pi}}e^{\frac{-(d - \mu)^{2}}{2\sigma^{2}}}, 
\end{equation}
where $d$ indicates the Euclidean distance between a pixel and the center of the distortion block. In our implementation, $\mu$ and $\sigma$ are set to 0 and 1 respectively.

\subsubsection{Network Architecture}
\ \

\noindent \textbf{Global Perception Module:} For the global perception module, its architecture can be divided into two sub-networks, a backbone network, and a header network. Specifically, the general representation of the global distortion context is extracted using the backbone network composed of convolutional layers. This representation indicates the high-level information including the semantic features. Any prevalent networks such as VGG16 \cite{ref30}, ResNet \cite{ref31}, and InceptionV3 \cite{ref53} (without fully connected layers) can be plugged into the backbone network. We pretrain the backbone network on ImageNet \cite{ref33} and fine-tune on our synthesized distorted image dataset. The header network contains three fully connected layers. It aggregates the input's general representation and further abstracts the high-level information in the form of a feature vector, which . The numbers of units for these layers are 4096, 2048, and 1024. The activation functions for all of the fully connected layers are ReLUs. The extracted features of the global distortion context, dubbed as $\mathcal{F}_g$, are combined with the features of the local distortion context, derived from the local Siamese module.

\noindent \textbf{Local Siamese Module:} The local Siamese module consists of $n$ components, each component also can be divided into a backbone network and a header network. In detail, we first use two convolutional layers to extract the low-level features from the input local distortion context. Then, we feed the feature maps into a pyramid residual module consisting of five residual blocks and get the high-level features. The pyramid residual module shares the weights in each component. Subsequently, a header network with three fully connected layers aggregates the general representation of the local distortion features: $[\mathcal{F}_l^{(1)} \ \ \mathcal{F}_l^{(2)} \ \ \cdots \ \ \mathcal{F}_l^{(n)}]$, which are one-to-one correspondence to the estimated ordinal distortion $\hat{\mathcal{D}} = [\hat \delta(r_1) \ \ \hat \delta(r_2) \ \  \cdots \ \ \hat \delta(r_n)]$.

Having observed that the distortion degree increases with the distance of a pixel to the principal point, we design a distortion-aware perception layer to extract the different distortion features. The motivation is illustrated in Fig. \ref{Fig:da}. In general, the filter size indicates the size of the receptive field, which determines the context of reasoning features. Therefore, it is more reasonable to grasp the prior knowledge of the variable distortion distribution using filters with different sizes instead of the same size. In our implementation, for different distortion blocks of a patch along the extended path (the white arrows in Fig. \ref{Fig:da}), we use convolutional filters with increasing sizes to extract the distortion features. Concretely, the distortion-aware perception layer is applied before feeding the input contexts to the network. For the local distortion context, the distortion blocks $\Theta = [\theta_1 \ \ \theta_2  \ \ \cdots \ \ \theta_n]$ are processed using the filters with sizes of $W_{l1}\times H_{l1}, W_{l2}\times H_{l2}, \cdots, W_{ln}\times H_{ln}$, from small to large. Namely, all sizes of filters satisfies the following relationship: $W_{l1}\times H_{l1} < W_{l2}\times H_{l2} < \cdots < W_{ln}\times H_{ln}$. As a result, our learning model can explicitly perceive the different degrees of distortions in a distorted image, thus achieving a better approximation of ordinal distortion. The relevant experimental results will be described in Section \ref{sec43}.

\noindent \textbf{Distortion Estimation Module:}
To comprehensively reason the distortion information, we combine each local distortion feature with the global distortion feature and fuse these features using two fully connected layers $f$, which constructs a hybrid feature vector $\mathcal{F}_h$:
\begin{equation}\label{eq-fusion}
\mathcal{F}_h = f([\mathcal{F}_l^{(1)} \oplus \mathcal{F}_g \ \ \mathcal{F}_l^{(2)} \oplus \mathcal{F}_g \ \ \cdots \ \ \mathcal{F}_l^{(n)} \oplus \mathcal{F}_g ]),
\end{equation}
where $\oplus$ represents the concatenation operation. Finally, a fully connected layer $F$ with the unit number of $n$ and linear activation function takes the $\mathcal{F}_h$ as input, estimating the ordinal distortion $\hat{\mathcal{D}}$ of a distorted image by

\begin{equation}\label{eq-estimation}
\hat{\mathcal{D}} = [\hat \delta(r_1) \ \ \hat \delta(r_2) \ \  \cdots \ \ \hat \delta(r_n)] = F(\mathcal{F}_h).
\end{equation}

\subsubsection{Training Loss}
After predicting the distortion labels of a distorted image, it is direct to use the distance metric loss such as $\mathcal{L}_1$ loss or $\mathcal{L}_2$ loss to learn the network parameters. Nevertheless, these loss functions cannot measure the ordered relationship in the distortion labels, while the proposed ordinal distortion possesses a strong ordinal correlation in terms of the distortion distribution. To this end, we regard the distortion estimation problem as an ordinal distortion regression problem and design an ordinal distortion loss to train our learning model.

Suppose that the ground truth ordinal distortion $\mathcal{D} = [\delta(r_1) \ \ \delta(r_2) \ \  \cdots \ \ \delta(r_n)]$ is an increasing vector, which means $\delta(r_1) < \delta(r_2) < \cdots < \delta(r_n)$. Due to the available distortion parameters in dataset, we can easily get the ground truth of ordinal distortion of any single image based on Eq. \ref{eq5}. Recall that $\mathcal{F}_g$ indicates the global distortion feature which is extracted by the global perception module $M_{gp}$; $[\mathcal{F}_l^{(1)} \ \ \mathcal{F}_l^{(2)} \ \ \cdots \ \ \mathcal{F}_l^{(n)}]$ indicate the local distortion features which are extracted by the local Siamese module $M_{ls}$. Subsequently, a distortion estimation module $M_{de}$ fuses the global feature and local features into a hybrid feature vector $\mathcal{F}_h$, that is used to predict the target ordinal distortion $\hat{\mathcal{D}} = [\hat \delta(r_1) \ \ \hat \delta(r_2) \ \  \cdots \ \ \hat \delta(r_n)]$. Let $\xi$ contains the weights of the final fully connected layer $F$, and then the ordinal distortion loss $\mathcal{L}(\mathcal{F}_h, \xi)$ can be described by the following formulation over the entire sequence:

\begin{equation}\label{eq-loss}
\mathcal{L}(\mathcal{F}_h, \xi) = \frac{1}{n}\sum_{i=1}^{n}{(1 + \mathcal{C}_{o})\mathcal{L}_d(i, \mathcal{F}_h, \xi)}.
\end{equation}

The term $\mathcal{C}_{o}$ is to weight the loss function and measures the ordinal correlation in $\hat{\mathcal{D}}$:

\begin{equation}\label{eq-loss1}
\mathcal{C}_{o} = \sum_{k=1}^{i}{\log(\mathcal{P}_i^k)} + \sum_{k=i+1}^{n}{\log(1 - \mathcal{P}_i^k)},
\end{equation}
where $\mathcal{P}_i^k = P(\hat{\delta}(r_i) > \hat{\delta}(r_k))$ indicates the probability that $\hat{\delta}(r_i)$ is larger than $\hat{\delta}(r_k)$. $\mathcal{L}_d(i, \mathcal{F}_h, \xi)$ minimizes the difference between the $\hat{\mathcal{D}}$ and the ground truth ${\mathcal{D}}$ based on the smooth $\mathcal{L}_1$ measurement \cite{61}:
\begin{eqnarray}\label{eq-loss2}
\mathcal{L}_d(i, \mathcal{F}_h, \xi)=
\left\{
\begin{array}{lll}
0.5\Phi_i^2, \ \ \ \ \ \ &if \ \ |\Phi_i| \le 1. \\
|\Phi_i| - 0.5, &otherwise,
\end{array}
\right.
\end{eqnarray}
where $\Phi_i = \hat \delta(r_i) - \delta(r_i)$. The $\mathcal{L}_d$ with smooth $\mathcal{L}_1$ measurement can be cast as the composition of the $\mathcal{L}_1$ and $\mathcal{L}_2$ losses, which can eliminate the exploding gradient problem during the training process. Therefore, our ordinal distortion loss function reasons both the increasing ordinal correlation in the predicted elements and the accurate distortion levels.

\subsection{Ordinal Distortion to Distortion Parameter}
\label{s33}

Once the ordinal distortion is estimated by neural networks, the distortion coefficients $\mathcal{K} = [k_1 \ \ k_2 \ \ \cdots \ \ k_n]$ of a distorted image can be easily obtained by
\begin{equation}\label{eq8}
\begin{bmatrix}
k_1 \ \ k_2 \ \ \cdots \ \ k_n
\end{bmatrix} ={\begin{bmatrix}
\delta(r_1) - 1\\\\
\delta(r_2) - 1\\\\
\vdots  \\ \\
\delta(r_n) - 1\\
\end{bmatrix}}^{\rm T}  {\begin{bmatrix}
r_1^2  &  r_2^2  & \cdots\ &r_n^2\\ \\
r_1^4  &  r_2^4  & \cdots\ &r_n^4\\ \\
 \vdots   & \vdots & \ddots  & \vdots  \\ \\
r_1^{2n}  &  r_2^{2n}  & \cdots\ &r_n^{2n}\\
\end{bmatrix}}^{-1}_{.}
\end{equation}
For clarity, we rewrite Eq. \ref{eq8} as follows:
\begin{equation}\label{eq9}
\mathcal{K} = \mathcal{D}^{*} \cdot \mathcal{R}^{-1},
\end{equation}
where $\mathcal{D}^{*} = \hat{\mathcal{D}} - [\underbrace{1 \ \ 1 \ \ \cdots \ \ 1}_{n}]$ and $\hat{\mathcal{D}}$ expresses the estimated ordinal distortion, and the location information with different powers is included in $\mathcal{R}$.

Finally, the rectified image can be warped by each pixel of the distorted image using the computed distortion parameters based on Eq. \ref{eq1} or Eq. \ref{eq3}.

\begin{algorithm}[t]
	\renewcommand{\algorithmicrequire}{\textbf{Input:}}
	\renewcommand{\algorithmicensure}{\textbf{Output:}}
	\caption{Training Process of The Proposed Network}
	\label{alg:1}
	\begin{algorithmic}[1]
	\REQUIRE Distorted Image ${I^d}$
	\ENSURE Ordinal Distortion $\hat{\mathcal{D}} = [\hat{\delta(r_1)} \ \ \hat{\delta(r_2)} \ \ \cdots \ \ \hat{\delta(r_n)}]$
	
	\STATE Crop and flip $I^d$ into four distortion elements $\Pi = [\pi_1 \ \ \pi_2 \ \ \pi_3 \ \ \pi_4]$ 
	\REPEAT
	\FORALL{$\pi_i \in \Pi$}
	\STATE Generate the feature vector of overall distortion $\mathcal{F}_g^{(i)}$ using the global perception module $M_{gp}$:  $\mathcal{F}_g^{(i)} \leftarrow M_{gp}(\pi_i)$
	\STATE  Crop $\pi_i$ into same size of distortion blocks $\Theta^{(i)} = [\theta_1^{(i)} \ \ \theta_2^{(i)} \ \ \cdots \ \ \theta_n^{(i)}]$
	\FORALL{$\theta_j^{(i)} \in \Theta^{(i)}$}
	\STATE Generate the feature vector of detailed distortion $\mathcal{F}_l^{(i, j)}$ using the local Siamese module  $M_{ls}$:  $\mathcal{F}_l^{(i, j)} \leftarrow M_{ls}(\theta_j^{(i)})$
	\ENDFOR
	\STATE Generate the hybrid feature vector $\mathcal{F}_h^{(i)}$ by fusing $\mathcal{F}_g^{(i)}$ and $[\mathcal{F}_l^{(i, 1)} \ \ \mathcal{F}_l^{(i, 2)} \ \ \cdots \ \ \mathcal{F}_l^{(i, n)}]$ based on Eq. \ref{eq-fusion} 
	\STATE Estimate the $\hat{\mathcal{D}}$ using $\mathcal{F}_h^{(i)}$ based on Eq. \ref{eq-estimation}
	\STATE Update the parameters of neural network by optimizing the difference of $\hat{\mathcal{D}}$ and the ground truth ${\mathcal{D}}$ based on Eq. \ref{eq-loss}
	\ENDFOR
    
	\UNTIL{Convergence}
	\end{algorithmic}  
\end{algorithm}

In summary, we argue that by presenting our distortion rectification framework, we can have the following advantages.

1. The proposed ordinal distortion is a learning-friendly representation for neural networks, which is explicit and homogeneous compared with the implicit and heterogeneous distortion parameters. Thus, our learning model gains sufficient distortion perception of features and shows faster convergence. Moreover, this representation enables more efficient learning with less data required.

2. The local-global associate ordinal distortion estimation network considers different scales of distortion features, jointly reasoning the local distortion context and global distortion context. Also, the devised distortion-aware perception layer boosts the feature extraction of different degrees of distortion.

3. Our ordinal distortion loss fully measures the strong ordinal correlation in the proposed representation, facilitating the accurate approximation of distortion distribution.

4. We can easily calculate the distortion parameters with the estimated ordinal distortion. In contrast to previous methods, our method can handle various camera models and different distortion types due to the unified learning representation. 
    
\begin{algorithm}[t]
	\renewcommand{\algorithmicrequire}{\textbf{Input:}}
	\renewcommand{\algorithmicensure}{\textbf{Output:}}
	\caption{Test Process of The Proposed Network}
	\label{alg:2}
	\begin{algorithmic}[1]
	\REQUIRE Distorted Image ${I^d}$
	\ENSURE Rectified Image ${I^r}$
	
	\STATE Crop and flip $I^d$ into four distortion elements $\Pi = [\pi_1 \ \ \pi_2 \ \ \pi_3 \ \ \pi_4]$, randomly feed one element $\pi_t$ into the trained network

	\STATE Generate the feature vector of overall distortion $\mathcal{F}_g^{(t)}$ using the global perception module $M_{gp}$:  $\mathcal{F}_g^{(t)} \leftarrow M_{gp}(\pi_t)$
	\STATE  Crop $\pi_t$ into same size of distortion blocks $\Theta^{(t)} = [\theta_1^{(t)} \ \ \theta_2^{(t)} \ \ \cdots \ \ \theta_n^{(t)}]$
	\FORALL{$\theta_j^{(t)} \in \Theta^{(t)}$}
	\STATE Generate the feature vector of detailed distortion $\mathcal{F}_l^{(t, j)}$ using the local Siamese module  $M_{ls}$:  $\mathcal{F}_l^{(t, j)} \leftarrow M_{ls}(\theta_j^{(t)})$
	\ENDFOR
	\STATE Generate the hybrid feature vector $\mathcal{F}_h^{(t)}$ by fusing $\mathcal{F}_g^{(t)}$ and $[\mathcal{F}_l^{(t, 1)} \ \ \mathcal{F}_l^{(t, 2)} \ \ \cdots \ \ \mathcal{F}_l^{(t, n)}]$ based on Eq. \ref{eq-fusion} 
	\STATE Estimate the $\hat{\mathcal{D}}$ using $\mathcal{F}_h^{(t)}$ based on Eq. \ref{eq-estimation}
	\STATE Compute the distortion coefficients $\hat{\mathcal{K}}$ using the $\hat{\mathcal{D}}$ based on Eq. \ref{eq8}
	\STATE Warp each pixel of $I^d$ using $\hat{\mathcal{K}}$ based on Eq. \ref{eq3}, to obtain $I^r$

	\end{algorithmic}  
\end{algorithm}

\section{Experiments}
\label{sec4}
In this section, we first state the details of the synthetic distorted image dataset and the training process of our learning model. Subsequently, we analyze the learning representation for distortion estimation. To demonstrate the effectiveness of each module in our framework, we conduct an ablation study to show the different performances. Additionally, the experimental results of our approach compared with the state-of-the-art methods are exhibited, in both quantitative measurement and visual qualitative appearance. Finally, we discuss two main limitations of our approach and present the possible solutions for future work.

\subsection{Implementation Settings}
\label{sec41}
\noindent \textbf{Dataset:} We construct a standard synthetic distorted image dataset in terms of the division model discussed in Section \ref{s31}. The original images are collected from the MS-COCO dataset \cite{56}. Following the implementations of previous literature \cite{ref11, ref48, ref36}, we also use a $4^{th}$ order polynomial based on Eq. \ref{eq3}, which is able to approximate most projection models with high accuracy. Additionally, all of the distortion coefficients are randomly generated from their corresponding ranges: $k_1 \in [-e^{-3}, -e^{-8}]$, $k_2 \in [-e^{-7}, -e^{-12}]$ or $[e^{-12}, e^{-7}]$, $k_3 \in [-e^{-11}, -e^{-16}]$ or $[e^{-16}, e^{-11}]$, and $k_4 \in [-e^{-15}, -e^{-20}]$ or $[e^{-20}, e^{-15}]$. Our synthetic dataset contains 20,000 training images, 2,000 test images, and 2,000 validation images.\\
\textbf{Training/Testing Setting:} 
We train our learning model on a NVIDIA RTX 2080 Ti GPU for 300 epochs, and the mini-batch size is 128. The backbone network of the global perception module is pre-trained on the ImageNet \cite{ref33}, and we fine-tune the learning model using the constructed synthetic distorted image dataset with a relatively small learning rate $5\times10^{-4}$, following the principle of transfer learning. The Adam \cite{ref34} is chosen as the optimizer with the parameters $\beta_1 = 0.5$ and $\beta_2 = 0.9$.

\begin{figure}[t]
	\begin{center}
		\includegraphics[width=1\linewidth]{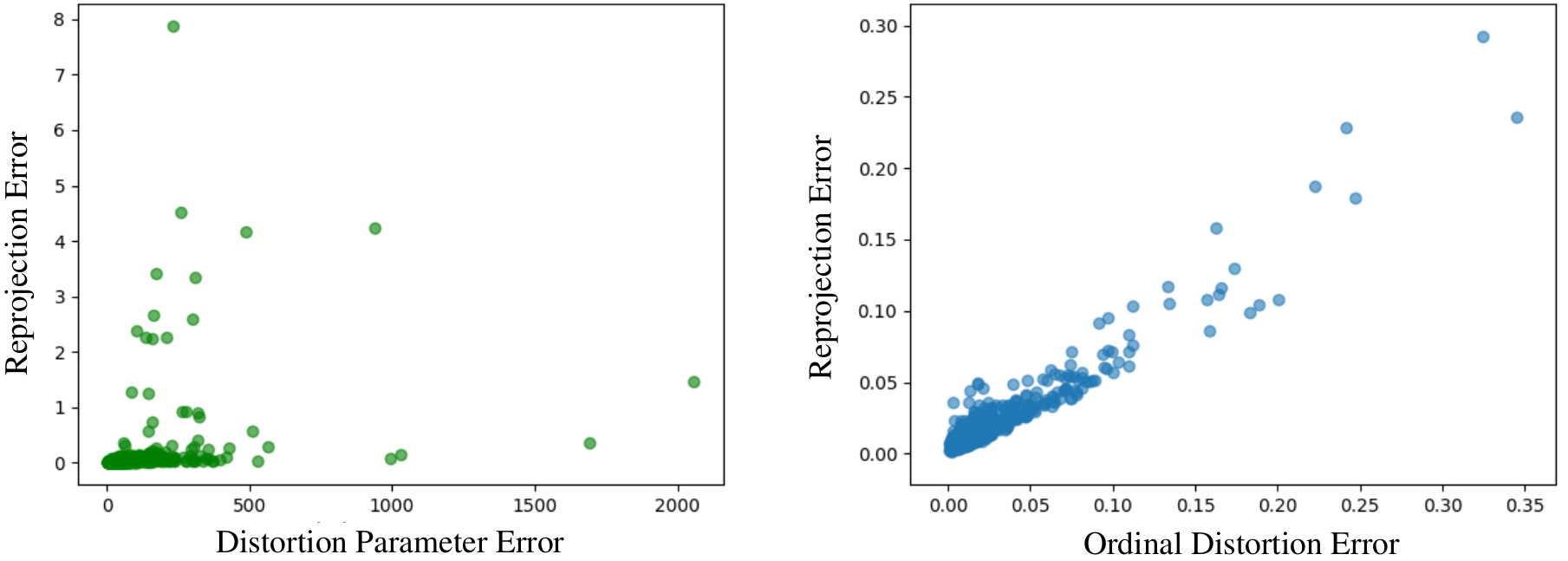}
		\caption{Comparison of two learning representations for distortion estimation, distortion parameter (left) and ordinal distortion (right). In contrast to the ambiguous relationship between the distortion distribution and distortion parameter, the proposed ordinal distortion displays an evident positive correlation to the distortion reprojection error.}
		\label{Fig:dl_dp}
	\end{center}
	%\vspace{-0.5em}
\end{figure}

\begin{figure*}[t]
	\begin{center}
		\includegraphics[width=1\linewidth]{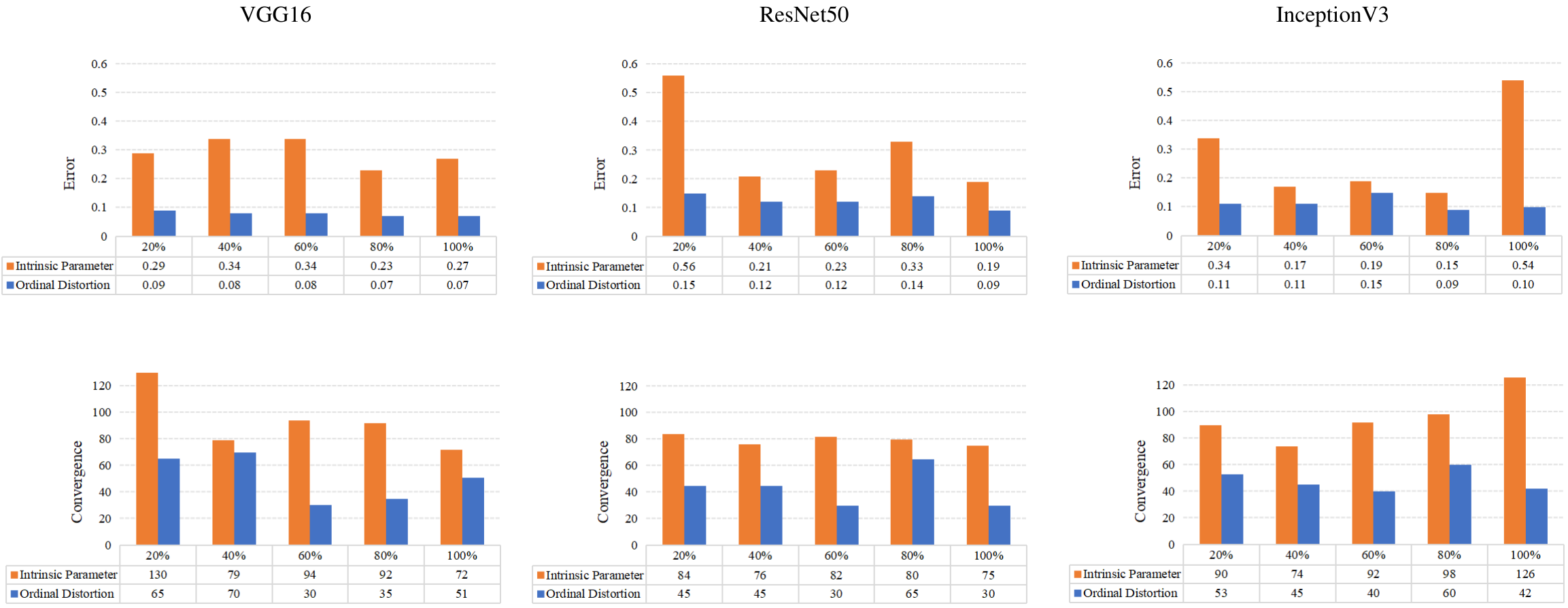}
		\caption{Analysis of two learning representations in terms of the error and convergence. We show the the histogram of error (top) and convergence (bottom) of two learning representations using three backbone networks, VGG16, ResNet50, and InceptionV3. Compared with the distortion estimation task, our proposed ordinal distortion estimation task achieves lower errors and faster convergence on all backbone networks.}
		\label{Fig:error_conv}
	\end{center}
	%\vspace{-0.5em}
\end{figure*}

\begin{figure*}[t]
	\begin{center}
		\includegraphics[width=1\linewidth]{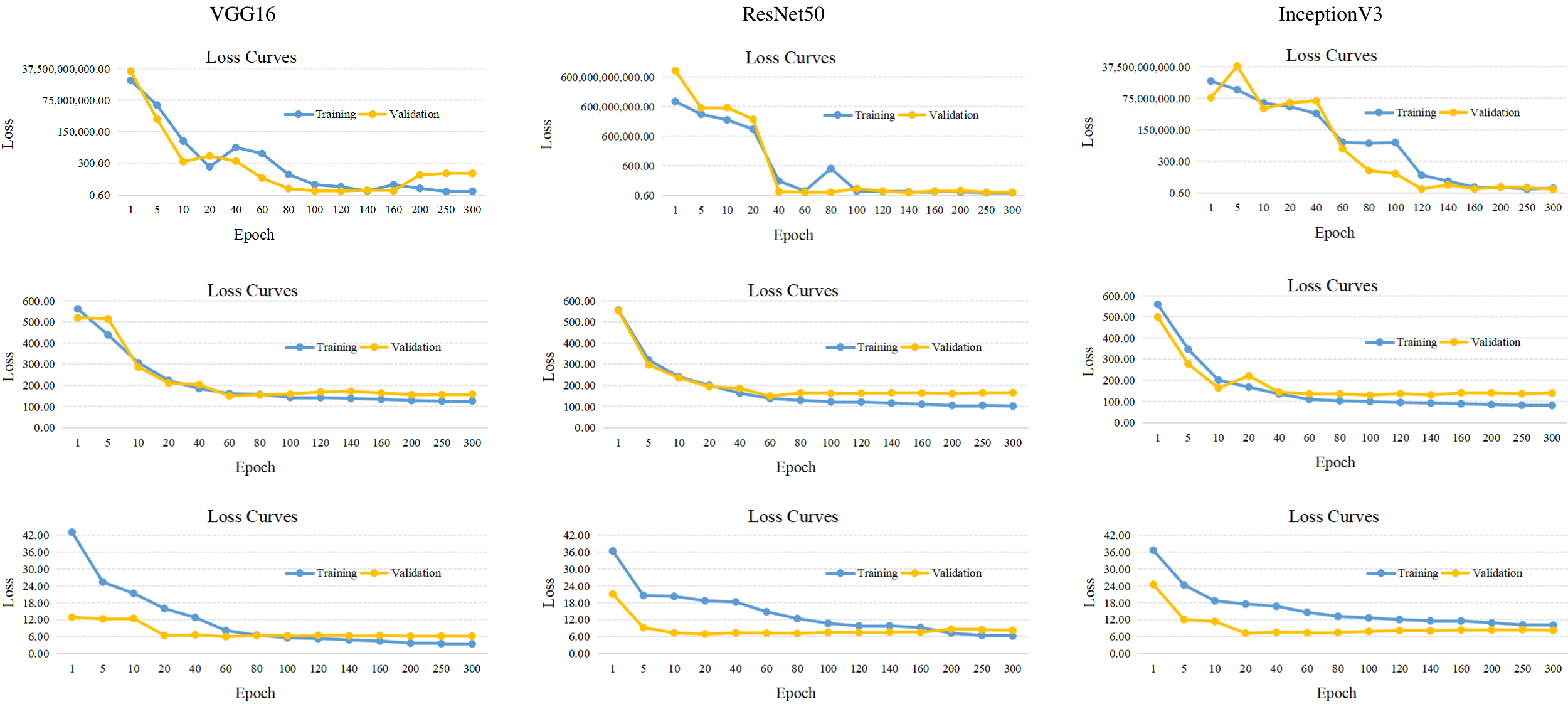}
		\caption{Analysis of two learning representation in terms of the training and validation loss curves. We show the learning performance of the distortion parameter estimation without (top) and with (middle) the normalization of magnitude, and the ordinal distortion estimation (bottom). Our proposed ordinal distortion estimation task displays the fast convergence and stable trend on both training and validation sets.}
		\label{Fig:loss}
	\end{center}
	%\vspace{-0.5em}
\end{figure*}

In the training stage, we crop each distorted image into four distortion elements and learn the parameters of the neural network using all data. Note that this training process is data-independent, where each part of the entire image is fed into the network one by one without the data correlation. In the test stage, we only need one distortion element, i.e., 1/4 of an image, to estimate the ordinal distortion. For a clear exhibition of our approach, we draw the detailed algorithm schemes of the training process and test process as listed in Algorithm \ref{alg:1} and Algorithm \ref{alg:2}, respectively.\\
\textbf{Evaluation Metrics:} Crucially, evaluating the performance of different methods with reasonable metrics benefits experimental comparisons. In the distortion rectification problem, the corrected image can be evaluated with the peak signal-to-noise ratio (PSNR) and the structural similarity index (SSIM). For the evaluation of the estimated distortion label, it is straightforward to employ the root mean square error (RMSE) between the estimated coefficients $\hat{\mathcal{K}}$ and ground truth coefficients $\mathcal{K}$:
\begin{equation}\label{eq12}
RMSE = \frac{1}{N}\sum_{i=1}^N\sqrt{{(\hat{\mathcal{K}_i} - \mathcal{K}_i)}^2},
\end{equation}
where $N$ is the number of estimated distortion coefficients. However, we found that different groups of distortion coefficients may display similar distortion distributions in images. To more reasonably evaluate the estimated distortion labels, we propose a metric based on the reprojection error, mean distortion level deviation (MDLD):
\begin{equation}\label{eq13}
MDLD = \frac{1}{WH}\sum_{i=1}^W\sum_{j=1}^H{|\hat{\delta}(i, j) - \delta(i, j)|},
\end{equation}
where $W$ and $H$ are the width and height of a distorted image, respectively. The ground truth distortion level $\delta(i, j)$ of each pixel can be obtained using Eq. \ref{eq5}.

In contrast to RMSE, MDLD is more suitable for parameter evaluation due to the uniqueness of the distortion distribution. Moreover, RMSE fails to evaluate the different numbers and attributes of estimated parameters for different camera models. Thanks to the objective description of the distortion, MDLD is capable of evaluating different distortion estimation methods using different camera models.

\begin{table}
  \caption{The learning-friendly rates of two learning representation evaluated with three backbone networks.}
  \label{tab:1}
	 \centering
   \begin{tabular}{p{2.6cm}<{\centering}p{1.3cm}<{\centering}p{1.3cm}<{\centering}p{1.5cm}<{\centering}}
    \toprule
    Learning Representation & VGG16 & ResNet50 & InceptionV3 \\
    \hline
    \midrule
    Distortion Parameter & 0.50 & 0.60 & 0.59\\
    Ordinal Distortion & \textbf{2.23} & 1.43 & 1.50\\
  \bottomrule
\end{tabular}
\end{table}

\begin{figure*}
	\begin{center}
		\includegraphics[width=1\linewidth]{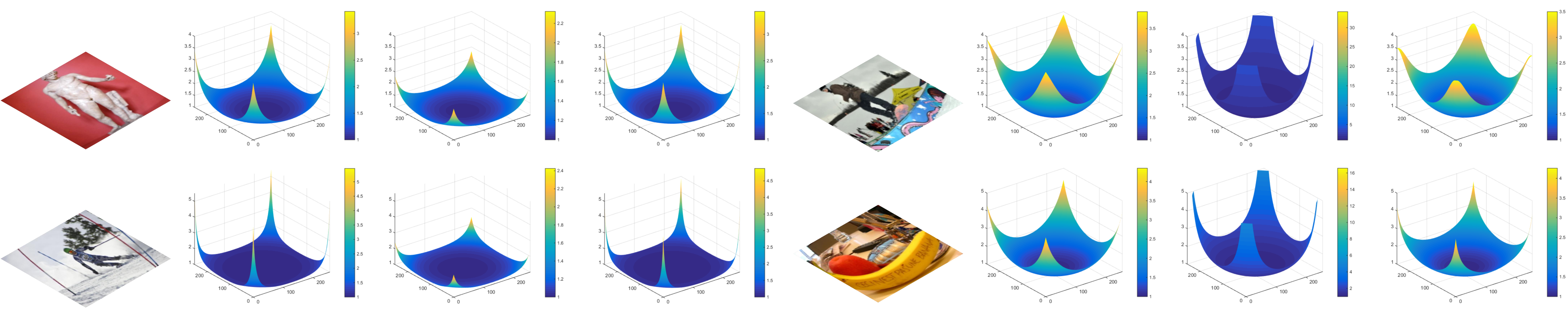}
		\caption{Qualitative comparison of two learning representations. For each comparison, we show the distorted image, the ground truth 3D DDM, the 3D DDM constructed by the estimated distortion parameter, and ordinal distortion, from left to right.}
		\label{Fig:ddm}
	\end{center}
	%\vspace{-0.5em}
\end{figure*}

\begin{figure}
	\begin{center}
		\includegraphics[width=1\linewidth]{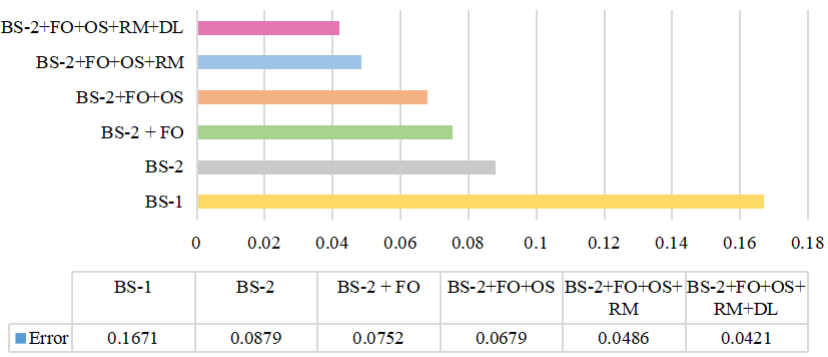}
		\caption{Ablation study of the proposed ordinal distortion estimation approach.}
		\label{Fig:ab}
	\end{center}
	%\vspace{-0.5em}
\end{figure}

\begin{figure}
	\begin{center}
		\includegraphics[width=1\linewidth]{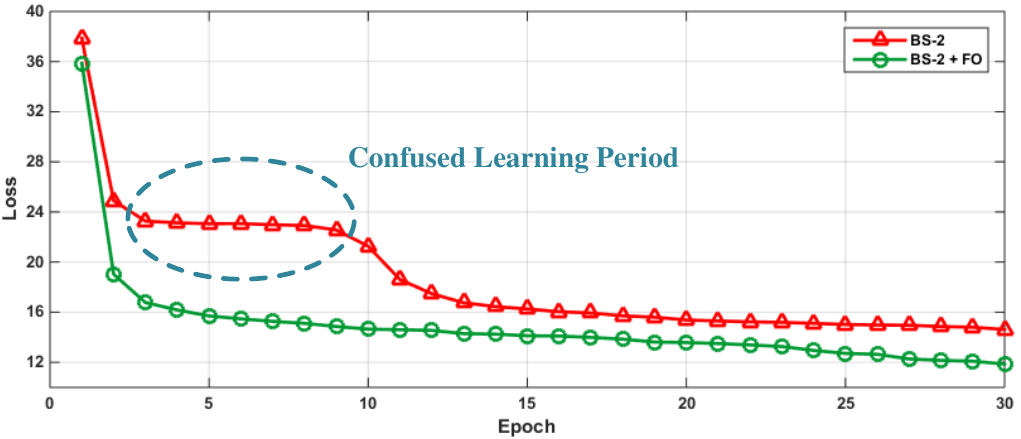}
		\caption{Training loss of first 30 epochs derived from the BS-2 and BS-2 + FO. The flip operation that normalizes the distortion distribution of inputs is able to significantly accelerate the convergence of the learning process.}
		\label{Fig:ab_loss}
	\end{center}
	%\vspace{-0.5em}
\end{figure}

\subsection{Analysis of Learning Representation}
\label{sec42}
Previous learning methods directly regress the distortion parameters from a distorted image. However, such an implicit and heterogeneous representation confuses the distortion learning of neural networks and causes the insufficient distortion perception. To bridge the gap between image feature and calibration objective, we present a novel intermediate representation, i.e., ordinal distortion, which displays a learning-friendly attribute for learning models. For an intuitive and comprehensive analysis, we compare these two representations from the following three aspects.

\noindent \textbf{Relationship to Distortion Distribution:} We first emphasize the relationship between two learning representations and the realistic distortion distribution of a distorted image. In detail, we train a learning model to estimate the distortion parameters and the ordinal distortions separately, and the errors of estimated results are built in the relationship to the distortion reprojection error. As shown in Fig. \ref{Fig:dl_dp}, we visualize the scatter diagram of two learning representations using 1,000 test distorted images. For the distortion parameter, its relationship to the distortion distribution is ambiguous and the similar parameter errors are related to quite different reprojection errors, which indicates that optimizing the parameter error would confuse the learning of neural networks. In contrast, the ordinal distortion error displays an evident positive correlation to the distortion distribution error, and thus the learning model gains intuitive distortion perception. Therefore, the proposed representation helps to decrease the error of distortion estimation.

\noindent \textbf{Distortion Learning Evaluation:} Then, we introduce three key elements for evaluating the learning representation: training data, convergence, and error. Supposed that the settings such as the network architecture and optimizer are the same, a better learning representation can be described from the less the training data is, the faster convergence and the lower error are. For example, a student is able to achieve the highest test grade (the lowest error) with the fastest learning speed and the least homework, meaning that he grasps the best learning strategy compared with other students. In terms of the above description, we evaluate the distortion parameter and ordinal distortion as shown in Fig. \ref{Fig:error_conv} and Fig. \ref{Fig:loss}.

To exhibit the performance fairly, we employ three common network architectures VGG16, ResNet50, and InceptionV3 as the backbone networks of the learning model. The proposed MDLD metric is used to express the distortion estimation error due to its unique and fair measurement for distortion distribution. To be specific, we visualize the error and convergence epoch when estimating two representations under the same number of training data in Fig. \ref{Fig:error_conv}, which is sampled with 20\%, 40\%, 60\%, 80\%, and 100\% from the entire training data. Besides, the training and validation loss curves of two learning representations are shown in Fig. \ref{Fig:loss}, in which the distortion parameters are processed without (top) and with (middle) the normalization of magnitude. From these learning evaluations, we can observe:

(1) Overall, the ordinal distortion estimation significantly outperforms the distortion parameter estimation in both convergence and accuracy, even if the amount of training data is 20\% of that used to train the learning model. Note that we only use 1/4 distorted image to predict the ordinal distortion. As we pointed out earlier, the proposed ordinal distortion is explicit to the image feature and is observable from a distorted image; thus it boosts the neural networks' learning ability. On the other hand, the performance of the distortion parameter estimation drops as the amount of training data decreases. In contrast, our ordinal distortion estimation performs more consistently due to the homogeneity of the learning representation. 

(2) For each backbone network, the layer depths of VGG16, InceptionV3, and ResNet50 are 23, 159, and 168, respectively. These architectures represent the different extraction abilities of image features. As illustrated in Fig. \ref{Fig:error_conv}, the distortion parameter estimation achieves the lowest error (0.15) using InceptionV3 as the backbone under 80\% training data, which indicates its performance requires more complicated and high-level features extracted by deep networks. With the explicit relationship to image features, the ordinal distortion estimation achieves the lowest error (0.07) using the VGG16 as the backbone under 100\% training data. This promising performance indicates the ordinal distortion is a learning-friendly representation, which is easy to learn even using a very shallow network.

(3) From the loss curves in Fig. \ref{Fig:loss}, the ordinal distortion estimation achieves the fastest convergence and best performance on the validation dataset. It is also worth to note that the ordinal distortion estimation already performs well on the validation at the first twenty epochs, which verifies that this learning representation yields a favorable generalization for neural networks. In contrast, suffering from the heterogeneous representation, the learning process of distortion parameter estimation displays a slower convergence. Moreover, the training and validation loss curves show unstable descend processes when the distortion parameters are handled without the normalization of magnitude, demonstrating the distortion parameter estimation is very sensitive to the label balancing.  

We further present a \textit{learning-friendly rate} ($\Gamma_{lr}$) to evaluate the effectiveness of learning representation or strategy quantitatively. To our knowledge, this is the first evaluation metric to describe the effectiveness of learning representation for neural networks. As mentioned above, the required training data, convergence, and error can jointly describe a learning representation, and thus we formulate the learning-friendly rate as follows

\begin{equation}\label{eq_lr}
\Gamma_{lr} = \frac{1}{M}\sum_{i=1}^N{\frac{T_i}{T}(\frac{1}{E_i}\log(2 - \frac{C_i}{C}))},
\end{equation}
where $M$ is the number of split groups, $E_i$, $T_i$, and $C_i$ indicate the error, number of training data, the epoch of convergence of the $i$-th group, respectively. $T$ and $C$ indicate the total number of training data and total training epochs for the learning model. We compute the learning-friendly rates of two learning representations and list the quantitative results in Table \ref{tab:1}. The results show that our scheme outperforms the distortion parameter estimation on all backbone settings, and thus the proposed ordinal distortion is much suitable for the neural network as a learning representation.

\noindent \textbf{Qualitative Comparison:} To qualitatively show the performance of different learning representations, we visualize the 3D distortion distribution maps (3D DDM) derived from the ground truth and these two schemes in Fig. \ref{Fig:ddm}, in which each pixel value of the distortion distribution map represents the distortion level. Since the ordinal distortion estimation pays more attention to the realistic distortion perception and reasonable learning strategy, our scheme achieves results much closer to the ground truth 3D DDM. Due to implicit learning, the distortion parameter estimation generates inferior reconstructed results, such as the under-fitting (left) and over-fitting (right) on the global distribution approximation as shown in Fig. \ref{Fig:ddm}.

\begin{figure*}
	\begin{center}
		\includegraphics[width=1\linewidth]{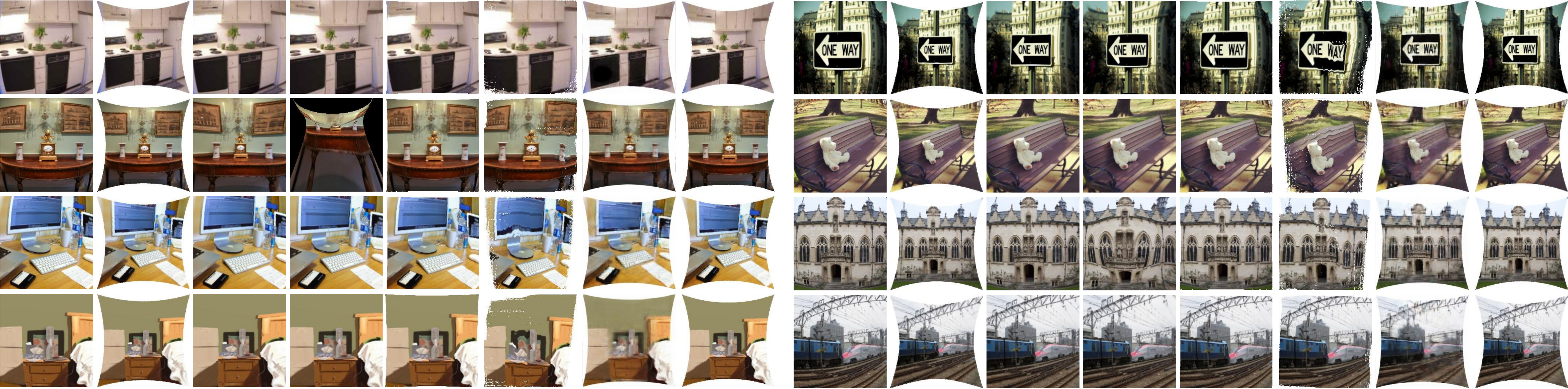}
		\caption{Qualitative evaluations of the rectified distorted images on indoor (left) and outdoor (right) scenes. For each evaluation, we show the distorted image, ground truth, and corrected results of the compared methods: Alem{\'a}n-Flores \cite{ref9}, Santana-Cedr{\'e}s \cite{ref45}, Rong \cite{ref10}, Li \cite{ref44}, and Liao \cite{ref46}, and rectified results of our proposed approach, from left to right.}
		\label{Fig:cp1}
	\end{center}
	%\vspace{-0.5em}
\end{figure*}

\begin{figure*}
	\begin{center}
		\includegraphics[width=1\linewidth]{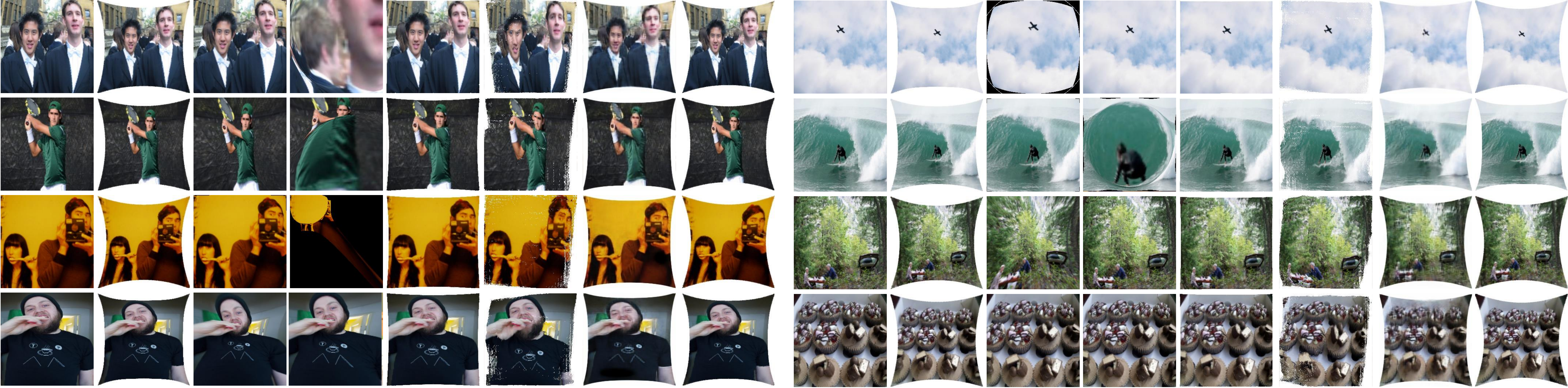}
		\caption{Qualitative evaluations of the rectified distorted images on people (left) and challenging (right) scenes. For each evaluation, we show the distorted image, ground truth, and corrected results of the compared methods: Alem{\'a}n-Flores \cite{ref9}, Santana-Cedr{\'e}s \cite{ref45}, Rong \cite{ref10}, Li \cite{ref44}, and Liao \cite{ref46}, and rectified results of our proposed approach, from left to right.}
		\label{Fig:cp2}
	\end{center}
	%\vspace{-0.5em}
\end{figure*}

\subsection{Ablation Study}
\label{sec43}
To validate the effectiveness of each component in our approach, we conduct an ablation study to evaluate the error of distortion estimation, as shown in Fig. \ref{Fig:ab}. Concretely, we first use the VGG16 network without the fully connected layers as the backbone of the ordinal distortion estimation network, based on the analysis of the learning representation in Section \ref{sec42}. Subsequently, we implement the learning model without the flip operation (FO) on global distortion context, ordinal supervision (OS), region of interest mask (RM), and distortion-aware perception layer (DL) as the baseline (BS), and then gradually add these removed components to show the different estimation performance. In addition, we perform two loss functions: $\mathcal{L}_2$ and $\mathcal{L}_{sm}$ to optimize the baseline model, in which $\mathcal{L}_{sm}$ is the smooth $\mathcal{L}_1$ loss function \cite{ref54} that combines the attributes of $\mathcal{L}_1$ and $\mathcal{L}_2$. We name these two types of baseline models as BS-1 and BS-2. During the training process, we crop four patches from the distorted image and shuffle the orders of all input patches. Subsequently, the patches are fed into the learning model. In the test stage, we only use one patch of a distorted image to evaluate the model.

Overall, the completed framework achieves the lowest error of distortion estimation as shown in Fig. \ref{Fig:ab}, verifying the effectiveness of our proposed approach. For the optimization strategy, the BS-2 used $\mathcal{L}_{sm}$ performs much better than BS-1 used $\mathcal{L}_{2}$ since the $\mathcal{L}_{sm}$ loss function boosts a more stable training process. Due to the effective normalization of distortion distribution, the network gains explicit spatial guidance with the flip operation on the global distortion context. We also show the training loss of the first 30 epochs derived from the BS-2 and BS-2 + FO in Fig. \ref{Fig:ab_loss}, where we can observe that the distribution normalization can significantly accelerate the convergence of the training process. On contrary, the BS-2 without flip operation suffers from a \textit{confused learning period} especially in the first 10 epochs, which indicates that the neural network is unsure how to find a direct optimization way from the distribution difference. Moreover, the ordinal supervision fully measures the strong ordinal correlation in the proposed representation, and thus facilitates the accurate approximation of distortion distribution. With the special attention mechanism and distortion feature extraction, our learning model gains further improvements using the region of interest mask and distortion-aware perception layer.

\subsection{Comparison Results}
\label{sec44}
In this part, we compare our approach with the state-of-the-art methods in both quantitative and qualitative evaluations, in which the compared methods can be classified into traditional methods \cite{ref9}\cite{ref45} and learning methods \cite{ref10}\cite{ref44}\cite{ref46}. Note that our approach only requires a patch of the input distorted image to estimate the ordinal distortion.

\begin{figure*}
	\begin{center}
		\includegraphics[width=1\linewidth]{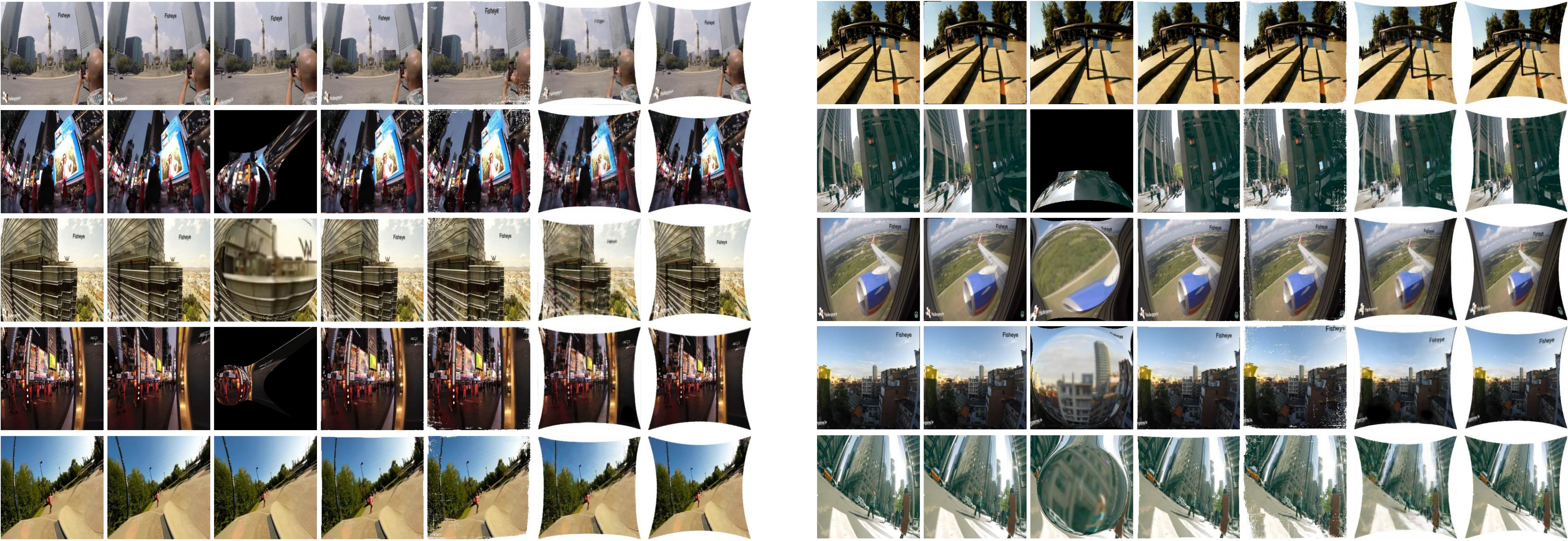}
		\caption{Qualitative evaluations of the rectified distorted images on real-world scenes. For each evaluation, we show the distorted image and corrected results of the compared methods: Alem{\'a}n-Flores \cite{ref9}, Santana-Cedr{\'e}s \cite{ref45}, Rong \cite{ref10}, Li \cite{ref44}, and Liao \cite{ref46}, and rectified results of our proposed approach, from left to right.}
		\label{Fig:cp3}
	\end{center}
	%\vspace{-0.5em}
\end{figure*}

\noindent \textbf{Quantitative Evaluation:}
To demonstrate a quantitative comparison with the state-of-the-art approaches, we evaluate the rectified images based on the PSNR (peak signal-to-noise ratio), SSIM (structural similarity index), and the proposed MDLD (mean distortion level deviation). All the comparison methods are used to conduct the distortion rectification on the test dataset including 2,000 distorted images. For the PSNR and SSIM, we compute these two metrics using the pixel difference between each rectified image and the ground truth image. For the MDLD, we first exploit the estimated distortion parameters to obtain all distortion levels of the test distorted image based on Eq. \ref{eq5}. Then, the value of MDLD can be calculated by the difference between estimated distortion levels and the ground truth distortion levels based on Eq. \ref{eq13}. Note that the generated-based methods such as Li \cite{ref44} and Liao \cite{ref46} directly learn the transformation manner of the pixel mapping instead of estimating the distortion parameters, so we only evaluate these two methods in terms of the PSNR and SSIM.

As listed in Table \ref{table:2}, our approach significantly outperforms the compared approaches in all metrics, including the highest metrics on PSNR and SSIM, as well as the lowest metric on MDLD. Specifically, compared with the traditional methods \cite{ref9, ref45} based on the hand-crafted features, our approach overcomes the scene limitation and simple camera model assumption, showing more promising generality and flexibility. Compared with the learning distortion rectification methods \cite{ref10}\cite{ref44}\cite{ref46}, which omit the prior knowledge of the distortion, our approach transfers the heterogeneous estimation problem into a homogeneous one, eliminating the implicit relationship between image features and predicted values in a more explicit expression. As benefits of the effective ordinal supervision and guidance of distortion information during the learning process, our approach outperforms Liao \cite{ref46} by a significant margin, with approximately 23\% improvement on PSNR and 17\% improvement on SSIM. Besides the high quality of the rectified image, our approach can obtain the accurate distortion parameters of a distorted image, which is crucial for the subsequent tasks such as the camera calibration. However, the generation-based methods \cite{ref44}\cite{ref46} mainly focus on the pixel reconstruction of a rectified image and ignore the parameter estimation.

\begin{table}[t]
\begin{center}
\caption{Quantitative evaluation of the rectified results obtained by different methods.}
\label{table:2}
\begin{tabular}{lccc}
\hline\noalign{\smallskip}
Comparison Methods & PSNR $\uparrow$ & SSIM $\uparrow$ & MDLD $\downarrow$ \\
\noalign{\smallskip}
\hline
\hline
\noalign{\smallskip}
Traditional Methods  &  &  &\\
Alem{\'a}n-Flores \cite{ref9} & 9.47 & 0.31 & 0.26 \\
Santana-Cedr{\'e}s \cite{ref45} & 7.90 & 0.25 & 1.18 \\
\noalign{\smallskip}
\hline
\noalign{\smallskip}
Learning Methods &   & & \\
Rong \cite{ref10} & 10.37 & 0.29 & 0.23\\
Li \cite{ref44} & 13.87 & 0.64 & - \\
Liao \cite{ref46} & 20.28 & 0.72 & - \\
Ours & \textbf{24.82} & \textbf{0.84} & \textbf{0.04} \\
\hline
\end{tabular}
\end{center}
\end{table}
\setlength{\tabcolsep}{1.6pt}

\noindent \textbf{Qualitative Evaluation:}
We visually compare the corrected results from our approach with state-of-the-art methods using our synthetic test set and the real distorted images. To show the comprehensive rectification performance under different scenes, we split the test set into four types of scenes: indoor, outdoor, people, and challenging scenes. The indoor and outdoor scenes are shown in Fig. \ref{Fig:cp1}, and the people and challenging scenes are shown in Fig. \ref{Fig:cp2}. Our approach performs well on all scenes, while the traditional methods \cite{ref9, ref45} show inferior corrected results under the scene that lacks sufficient hand-crafted features, especially in the people and challenging scenes. On the other hand, the learning methods \cite{ref10, ref44, ref46} lag behind in the sufficient distortion perception and cannot easily adapt to scenes with strong geometric distortion. For example, the results obtained by Rong \cite{ref10} show coarse rectified structures, which are induced by the implicit learning of distortion and simple model assumption. Li \cite{ref44} leveraged the estimated distortion flow to generate the rectified images. However, the accuracy of the pixel-wise reconstruction heavily relies on the performance of scene analysis, leading to some stronger distortion results under complex scenes. Although Liao \cite{ref46} generated better rectified images than the above learning methods in terms of global distribution; the results display unpleasant blur local appearances due to the used adversarial learning manner. In contrast, our results achieve the best performance on global distribution and local appearance, which benefit from the proposed learning-friendly representation and the effective learning model.

The comparison results of the real distorted image are shown in Fig. \ref{Fig:cp3}. We collect the real distorted images from the videos on YouTube, captured by popular fisheye lenses, such as the SAMSUNG 10mm F3, Rokinon 8mm Cine Lens, Opteka 6.5mm Lens, and GoPro. As illustrated in Fig. \ref{Fig:cp3}, our approach generates the best rectification results compared with the state-of-the-art methods, showing the appealing generalization ability for blind distortion rectification. To be specific, the salient objects such as buildings, streetlights, and roads are recovered into their original straight structures by our approach, which exhibits a more realistic geometric appearance than the results of other methods. Since our approach mainly focuses on the design of learning representation for distortion estimation, the neural networks gain more powerful learning ability to the distortion perception and achieve more accurate estimation results.

\subsection{Limitation Discussion}
In this work, we presented a new learning representation for the deep distortion rectification and implemented a standard and widely-used camera model to validate its effectiveness. The rectification results on the synthesized and real-world scenarios also demonstrated our approach's superiority compared with the state-of-the-art methods. Like most of the assumptions in the other works \cite{ref7, ref9, ref10, ref44, ref46, ref55}, our approach has two main limitations to extend to more complicated applications.

The first limitation is that the principal point needs to be at the center of the image. Observing that the principal point is slightly disturbed around the center of the image, we mainly consider the estimation of distortion coefficients using the proposed ordinal distortion in our work. Nevertheless, our method can be easily extended to more scenarios when the network predicts more target labels. For example, suppose we wish to estimate a principal point $(x_c, y_c)$ and four distortion coefficients $(k_1, k_2, k_3, k_4)$ of a distorted image (six variables in total). In that case, we only need to predict the ordinal distortion $\mathcal{D} = [\delta(r_1) \ \ \delta(r_2) \ \ \delta(r_3) \ \ \delta(r_4) \ \ \delta(r_5) \ \ \delta(r_6)]$ with two extra distortion levels $\delta(r_5)$ and $\delta(r_6)$ than the original scheme, namely, building simultaneous equations for solving six variables based on Eq. \ref{eq5}. Moreover, in our previous work \cite{62}, we developed a VGG-like network to regress the principal point given a distorted image, and then other distortion parameters are estimated accordingly. Thus, this sequential estimation solution also could be used in more complicated cases.

The second limitation is that the distortion needs to be radially symmetric. This problem may be addressed by the grid optimization technique in computer graphics, and we can teach the network to learn an asymmetric grid to warp each pixel of the distorted image. Based on the above limitations and the presented solutions, we plan to achieve a more comprehensive and robust distortion rectification framework in future work.

\section{Conclusion}
\label{sec5}
In this paper, we present a learning-friendly representation for the deep distortion rectification, bridging the gap between image feature and calibration objective. Compared with the implicit and heterogeneous distortion parameters, the proposed ordinal distortion offers three unique advantages: explicitness, homogeneity, and redundancy, enabling a sufficient and efficient learning on the distortion. To learn this representation, we design a local-global associated estimation network optimized with an ordinal distortion loss function, and a distortion-aware perception layer is used to boost the features extraction of different degrees of distortion. As the benefit of the proposed learning representation and learning model, our approach outperforms the state-of-the-art methods by a remarkable margin while only leveraging a part of data for distortion estimation.

\normalem
\bibliographystyle{ieeetr}
\bibliography{ref}
\end{document}